\documentclass[10pt,twocolumn,letterpaper]{article}

\usepackage{iccv}              % To produce the CAMERA-READY version
\definecolor{iccvblue}{rgb}{0.21,0.49,0.74}
\usepackage[pagebackref,breaklinks,colorlinks,allcolors=iccvblue]{hyperref}

\def\paperID{15480} % *** Enter the Paper ID here
\def\confName{ICCV}
\def\confYear{2025}

\usepackage{multirow}
\usepackage{kotex}
\usepackage{pifont}

\usepackage{soul}

\usepackage[capitalize]{cleveref}
\crefname{section}{Sec.}{Secs.}
\Crefname{section}{Section}{Sections}
\Crefname{table}{Table}{Tables}
\crefname{table}{Tab.}{Tabs.}
\Crefname{figure}{Figure}{Figures}
\crefname{figure}{Fig.}{Figs.}

\begin{document}

\title{SparseVoxFormer: Sparse Voxel-based Transformer for \\Multi-modal 3D Object Detection}

\author{Hyeongseok Son\textsuperscript{1} \quad Jia He\textsuperscript{2} \quad Seung-In Park\textsuperscript{1} \quad Ying Min\textsuperscript{2} \quad Yunhao Zhang\textsuperscript{2} \quad ByungIn Yoo\textsuperscript{1}\\
\textsuperscript{1}Samsung Electronics, AI Center, South Korea \\
\textsuperscript{2}Samsung R\&D Institute China Xi'an (SRCX), China\\
{\tt\small \{hs1.son, jia01.he, si14.park ying.min, yunhao.zhang, byungin.yoo\}@samsung.com}
% \and
% Jia He\\
% % First line of institution2 address\\
% % {\tt\small secondauthor@i2.org}
}
\maketitle

\newcommand{\change}[1]{{\color{red}#1}}
\newcommand{\supp}[1]{{\color{black}#1}}
\newcommand{\son}[1]{{\textcolor{magenta}{hyeongseok: #1}}}

% math macro
\newcommand{\SAMLOC}{\mathcal{S}}
\newcommand{\LEK}{\mathcal{K}}
\newcommand{\JNT}{\mathcal{J}}
\newcommand{\POSQ}{\mathcal{P}}
\newcommand{\CONQ}{\mathcal{E}}
\newcommand{\LOSS}{\mathcal{L}}

\newcommand{\noff}{N_p}
\newcommand{\nhead}{N_h}
\newcommand{\nscl}{N_s}

\newcommand{\REAL}{\mathbb{R}}
\newcommand{\PTS}{\mathbf{p}}

\begin{abstract}
Most previous 3D object detection methods that leverage the multi-modality of LiDAR and cameras utilize the Bird's Eye View (BEV) space for intermediate feature representation. However, this space uses a low x, y-resolution and sacrifices z-axis information to reduce the overall feature resolution, which may result in declined accuracy. To tackle the problem of using low-resolution features, this paper focuses on the sparse nature of LiDAR point cloud data. From our observation, the number of occupied cells in the 3D voxels constructed from a LiDAR data can be even fewer than the number of total cells in the BEV map, despite the voxels' significantly higher resolution. Based on this, we introduce a novel sparse voxel-based transformer network for 3D object detection, dubbed as SparseVoxFormer. Instead of performing BEV feature extraction, we directly leverage sparse voxel features as the input for a transformer-based detector. Moreover, with regard to the camera modality, we introduce an explicit modality fusion approach that involves projecting 3D voxel coordinates onto 2D images and collecting the corresponding image features. Thanks to these components, our approach can leverage geometrically richer multi-modal features while even reducing the computational cost. Beyond the proof-of-concept level, we further focus on facilitating better multi-modal fusion and flexible control over the number of sparse features. Finally, thorough experimental results demonstrate that utilizing a significantly smaller number of sparse features drastically reduces computational costs in a 3D object detector while enhancing both overall and long-range performance.
% Finally, we further sparsify  the SparseVoxFormer by employing multi-modal feature refinement and feature sparsification approaches specifically for sparse features. Thorough experimental results demonstrate that utilizing a significantly smaller number of sparse features drastically reduces computational costs in a 3D object detector while enhancing both overall and long-range performance.
\end{abstract}
\section{Introduction}
\label{sec:intro}
%% Sensor fusion is common approach in 3D object detection for autonomous driving. LiDAR-based approaches shows high performance, but weak for long range objects and cost problem of employing high-spec hardware. Recently image-based can be substitute. Multi-modal approach is a viable solution.
3D object detection is a critical task in real-world applications such as autonomous driving. A prominent approach to 3D object detection in autonomous driving involves using LiDAR sensors~\cite{yan2018second,shi2019point,lang2019pointpillars,shi2020pv,wang2020,li2021lidar}, primarily thanks to their ability to provide accurate localization. However, LiDAR sensors have clear limitations; the density of the point cloud significantly decreases as the distance from the sensor increases, leading to a considerable drop in accuracy for objects at long range~\cite{jiang2023semanticbevfusion}. Given that employing high-specification LiDAR sensors is cost-inefficient, a plausible solution would be to incorporate a camera modality.

Recent multi-modal approaches~\cite{bai2022transfusion,liang2022bevfusion,liu2023bevfusion,yang2022deepinteraction,fu2024,yin2024isfusion,song2024graphbev} combining LiDAR and camera data have achieved new state-of-the-art performances in 3D object detection for autonomous driving. These approaches typically use a BEV space to fuse multi-modal features from LiDAR and camera data, primarily due to the significant computational demands of directly utilizing high-resolution 3D features. Despite their practical achievements, these methods potentially lose 3D geometric information due to lower resolution and suppressed z-axis information. We believe that there is room for improvement by exploiting the rich geometric information present in 3D features.
% \son{Require more writing and good writing for the early part of introduction.}

%% Sparse nature
We observe that a point cloud of LiDAR data is inherently sparse, as are the voxels constructed from this data. Consequently, while the raw voxels occupy a high-resolution 3D space, the number of valid cells is not extensive. For instance, voxel features with a resolution of $360\times360\times11$ have a comparable number of cells to that of BEV features with a lower resolution of $180\times180$, resulting in 32,400 cells. This is surprising given that the voxel features originally have a total cell count 41 times greater than that of the BEV features. This suggests that by utilizing the sparsity of data, we can fully leverage the benefits of 3D voxel features and richer geometric information, achieving better performance while using fewer computational resources than when using BEV features.

%% Our approach
Based on this important observation, we propose a novel multi-modal 3D object detection framework directly using sparse voxel features, dubbed as SparseVoxFormer.
To this end, we employ a transformer decoder architecture like DETR~\cite{detr} for our detector because this structure can accept the sparse 3D voxel features intactly thanks to the nature of this architecture to receive 1D serialized inputs.
The conversion of dense voxel features into sparse representations can be achieved through straightforward filtering of zero-valued features. Obtaining voxel features with a higher resolution is done by deriving intermediate features from a conventional LiDAR backbone, which implying that we rather uses a less computational cost than that of BEV features.
% To supplement the potential of insufficient voxel encoding, we introduce a simple adaptation to voxel encoding, wherein the averaging of 3D points within each cell is enhanced by incorporating additional statistics such as standard deviation and point count to enable each voxel feature to encode fine-level geometric details. 

% As our new architecture exploits voxel features directly, it is important for these voxel features to carry rich 3D geometric features. However, current voxel construction approach may suppress the fine detail of points' distribution in each voxel due to simply averaging them. To remedy this problem, we present a modified simple voxel encoding approach, which containing an additional information such as the standard deviation and the number of points in each cell. 

Utilizing 3D voxel features has another benefit in multi-modal fusion with a image modality. 3D positional coordinates of all voxels are embedded to the voxel features so that each voxel feature can be explicitly and accurately projected to a corresponding image feature. By simply concatenating each voxel and the corresponding image feature, we can perform explicit and accurate multi-modal feature fusion. On the other hand, previous approaches depend on depth estimation for explicit fusion~\cite{liu2023bevfusion,yang2022deepinteraction,li2022unifying} or implicit fusion via transformer attention~\cite{yan2023cross}, which can eventually lead to erroneous multi-modal alignment and weak multi-modal information fusion.

%% advanced components, but too long...
% \son{Describing advanced components may make the introduction too long. How to solve it?}
% While our baseline model has already achieved comparable performance to that of previous state-of-the-art models, there is room for improvement by employing sparse feature-specific operations.
While our fundamental design has already successfully improved the performance of previous BEV-based models while significantly reducing the number of multi-modal features used, further enhancements can be achieved through the utilization of sparse feature-specific operations.
Firstly, the majority of parameters in a previous LiDAR backbone are used for processing dense features and are thus not utilized in our architecture, implying that our basic detector could use incompletely encoded features. To address this, we employ a recently proposed sparse feature refinement module~\cite{wang2023dsvt} to fully encode the geometric information in our sparse voxel features. We also discover that this refinement can be more effectively applied to our multi-modal fused features rather than solely to the LiDAR features.
Secondly, a problem arises due to the varying number of transformer tokens from LiDAR samples, which is caused by differing levels of sparsity. To address this, we implement an additional feature elimination technique. This not only reduces the computational cost but also maintains a consistent number of transformer tokens, irrespective of the sparsity variations in the LiDAR data.
% Current LiDAR backbones depends BEV feature extraction for geometric information encoding in the end, but in our case, we do not have this component. Therefore, our approach requires a more direct approach to construct voxel features with fine-level geometric information, which is appropriate for DSVT.

%Extensive experimental results demonstrate that our approach can effectively utilize sparse but fine-level geometric features possessed in 3D voxels with a higher resolution for 3D object detection. Finally, our approach achieves a state-of-the-art performance on the nuScenes dataset~\cite{caesar2020nuscenes} with a smaller computational cost. 
Extensive experimental results validate that sparse but accurately fused multi-modal features can effectively detect long-range objects by exploiting fine-level geometric features inherent in high-resolution 3D voxels. Finally, our approach achieves state-of-the-art performance on the nuScenes dataset \cite{caesar2020nuscenes} with a faster inference speed.

%While this innovative architecture have already achieved a comparable performance to the previous state-of-the-art performance with a smaller computational cost, we further explore Transformer architecture to fully exploit rich geometric information inherited from 3D voxel features with a higher resolution than that of BEV features.

To summarize, our contributions include:
\begin{itemize}
    \item A novel multi-modal 3D object detection framework, breaking the current paradigm to use a BEV space and achieving a state-of-the-art performance in 3D object detection on the nuScenes dataset,
    \item Direct utilization of sparse 3D voxel features to exploit a higher-resolution geometric information while reducing a computation cost,
    % \item A simple modification in voxel construction to make each raw cell possess fine-level geometric details,
    \item An accurate, explicit multi-modal fusion approach, realized by the benefit of 3D voxel features carrying 3D positional coordinates,
    \item Multi-modal feature refinement and additional feature sparsification for more sparse yet well-fused multi-modal features, which also enable flexible control over the number of sparse features.
    % \item A new state-of-the-art performance in 3D object detection on the Nuscenes dataset.
\end{itemize}

\section{Related Work}
\label{sec:related}
% \son{I am writing the related work. I got a main storyline but the details should be filled in more.}
% - 3D object detection의 경우, 대표적으로 KITTI, Waymo, Nuscenes 데이터셋들이 존재함, 본 논문에서는 이 중 360도를 multi-modal로 커버할 수 있는 Nucenes dataset을 중심으로 연구를 수행함
Typically, 3D object detection approaches for autonomous driving utilize datasets such as KITTI~\cite{Geiger2012CVPR}, Waymo Open Dataset~\cite{Sun_2020_CVPR}, and nuScenes~\cite{caesar2020nuscenes}. Given the differences in view coverage and multi-modality among these datasets, most approaches focus on one specific dataset. In this paper, we target multi-modal 3D object detection and thus specifically focus on the nuScenes dataset, which is unique in that it is the only one to provide 360\(^{\circ}\) view coverage and full multi-modality with LiDAR and camera sensors.
% \son{I am adding more references.}

%%%--------------------------------------------------------------------------------
\begin{figure*}[t]
    \centering
    \includegraphics[width=0.90\linewidth]{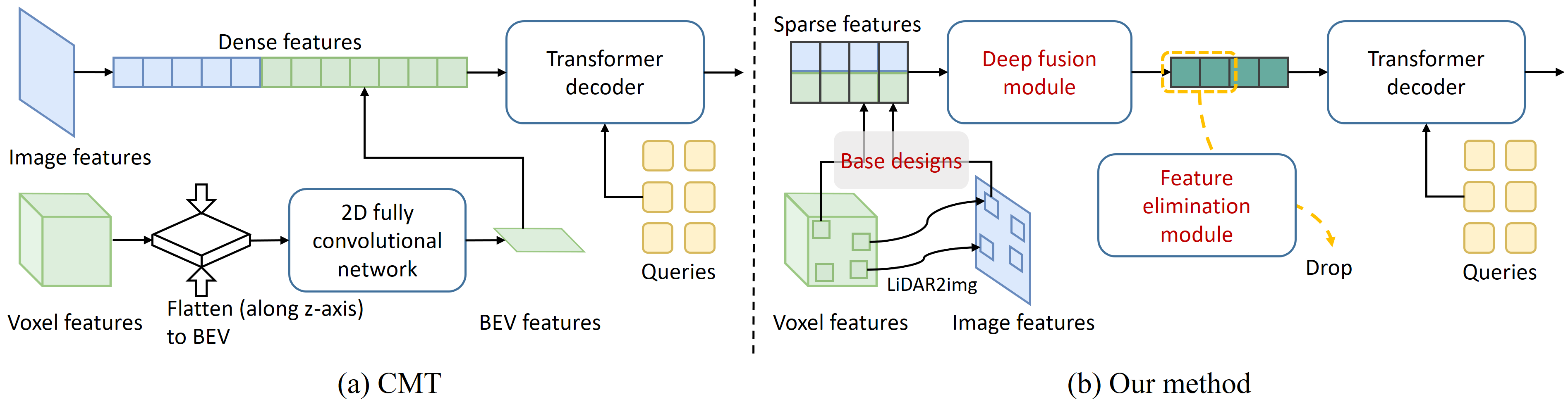}
    \vspace{-8pt}
    \caption{Architecture comparison between CMT~\cite{yan2023cross} and our SparseVoxformer. 
    % \son{Use this figure as a representative figure and we need to improve more.}    
    % \son{I am improving this figure.}
    }
    \vspace{-6pt}
    \label{fig:architecture}
\end{figure*}
%%%--------------------------------------------------------------------------------

% \vspace{-9pt}
\paragraph{3D object detection with a single modality}
3D object detection with a camera modality is a common setting in computer vision and has been extensively studied. In a standard 3D object detection task from a single image~\cite{qi2018frustum,brazil2019m3d,simonelli2019,wang21fcos3d,detr3d}, the input camera space could be sufficient for detection outputs. However, in autonomous driving, the output space needs to be a 3D world space surrounding the ego-vehicle, covered by multiple images with normal field-of-views. This introduces a representation gap between the input and output spaces. For this reason, several works~\cite{huang2021bevdet,li2023bevdepth} use a BEV space for their representation space. However, transforming image features to a 3D world space or BEV space requires additional depth information.

In the field of 3D object detection for autonomous driving, using a LiDAR modality~\cite{yan2018second,shi2019point,li2021lidar} is another mainstream area of research. Given that LiDAR data takes the form of point clouds, it provides high localization accuracy for objects. Unlike the camera modality, the LiDAR space can be considered as a unified space for both input and output modalities. However, 3D voxel features of LiDAR data require significant computational resources due to the number of cells. Hence, to reduce the computational complexity, recent works~\cite{lang2019pointpillars,shi2020pv,wang2020} also adopt a BEV space instead of a 3D space. As a result, recent approaches with either a LiDAR or camera~\cite{li2022bevformer,Yang2022BEVFormerVA} modality tend to use a BEV space for feature representation. However, this may result in overlooking 3D geometric information in high-resolution features, especially information along the z-axis.

More recently, Chen \etal\cite{chen2023voxenext} and Zhang \etal\cite{zhang2024safdnet} present frameworks for fully sparse 3D object detection based on a LiDAR modality, which uses sparse voxel features in 3D object detection heads. However, there are three significant distinctions from our approach. Firstly, they employ a sparse height compression module that suppresses the z-axis of the 3D voxel features into 2D sparse voxel features prior to inputting it into a 3D object detector. This can be regarded as using sparse BEV features. 
Secondly, they do not present a multi-modal setting incorporating a camera modality, whereas ours incorporates a simple yet effective multi-modal fusion approach suitable for sparse 3D voxel features. 
% Finally, considering only a LiDAR modality, our model outperforms their model, as will be demonstrated in Sec.~\ref{ssec:smaller_backbones}, due to our use of a Transformer-based detector for exploiting sparse 3D voxel features. This is also a novel concept that has not been explored in previous literature.
Finally, we employ a Transformer-based detector for exploiting sparse 3D voxel features, which is also a novel concept that has not been explored in previous literature.

%%%--------------------------------------------------------------------------------
\begin{figure*}[t]
\begin{center}
\includegraphics [width=0.90\linewidth] {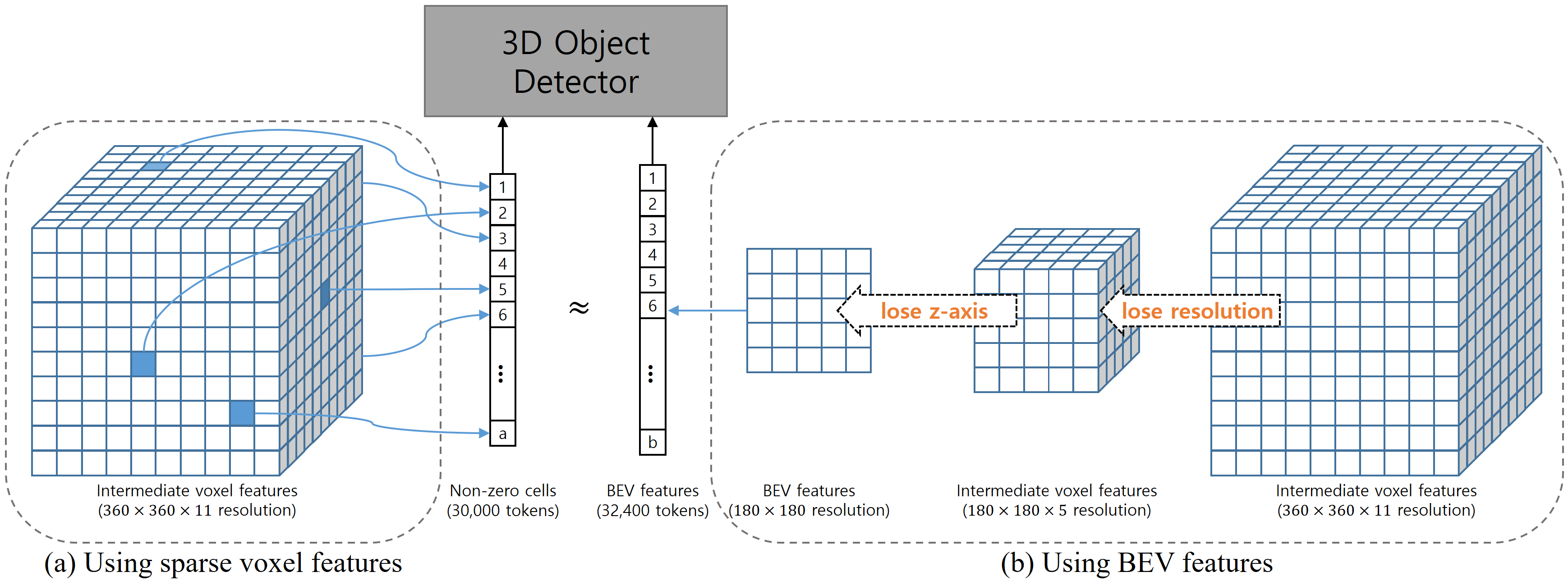}
\end{center}
\vspace{-10pt}
  \caption{A key idea to use sparse voxel features instead of BEV features. BEV features, obtained from lower-resolution and z-axis suppressed features, rather can produce a comparable number of tokens to that of higher-resolution voxel features.
  % \son{should be modified}
}
\label{fig:key_idea}
\vspace{-8pt}
\end{figure*}
%%%--------------------------------------------------------------------------------

\vspace{-6pt}
\paragraph{Multi-modal 3D object detection for autonomous driving}
% - Multi-modality
% 두 모달리티를 결합하려는 시도들이 있었고, 가장 좋은 성능을 거둠
% 대표적인 방법 BEVFusion, Deep interaction, Cross-modal transformer
% 여전히 LiDAR 처리를 위해 BEV space를 사용한다는 한계가 있고
% Multi-modal fusion을 위해 depth가 필요하거나, implicit interaction을 수행
% 본 방법은 3D coordinates가 보존되는 3D voxel-space을 이용하기 때문에 multi-modal fusion을 explicit하면서 정확하게 수행할 수 있음.

% 최근 UVTR에서 3D voxel-space 사용 -> 그러나 계산적으로 비효율적. 이미지 feature를 3D로 가져올 때 depth 활용 -> BEV-based 방법보다 낮은 성능
% 우리는 LiDAR의 sparse 특성을 잘 활용하여 3D voxel-space를 활용함에도 BEV-based 방법보다 계산량 측면에서 효율적이고 성능도 좋음

The fusion of LiDAR and camera sensors is a common approach in 3D object detection for autonomous driving, known for its synergistic relation. The camera modality provides semantic information, supplementing the sparsity at long-range distances inherent in the LiDAR modality. Conversely, the LiDAR modality supplies accurate localization information, which is often lacking in the camera modality.
However, merging these two types of data to form a unified feature set for multi-modal vision tasks can be challenging due to their inherent heterogeneity. To address this, most state-of-the-art approaches use a BEV space as a common ground for multi-modal fusion, following the developments in the literature regarding single modality processing.
Liang \etal\cite{liang2022bevfusion} and Liu \etal\cite{liu2023bevfusion} transform 2D image features into the BEV space by employing view transformation based on image depth estimation and then fuse them with LiDAR BEV features. Yang \etal~\cite{yang2022deepinteraction} introduce a cross-modal fusion approach that utilizes blocks for both LiDAR-to-camera and camera-to-LiDAR fusion. However, the multi-modal fusion in these approaches could be incomplete due to inaccuracies in depth distribution estimated from a single image. Additionally, the overhead of image view transformation could be significant.

% Fu \etal\cite{fu2024} and Song \etal\cite{song2024graphbev} address inaccurate multi-modal fusion, which occur by the nature of depth estimation-based image projection.
% Yin \etal\cite{yin2024isfusion} propose two fusion approaches, hierarchical multi-modal fusion in a BEV space and instance information fusion.
% However these methods still uses a BEV space for their multi-modal fusion, which may still induce a loss of 3D geometric information.
Fu \etal\cite{fu2024} and Song \etal\cite{song2024graphbev} address the issue of inaccurate multi-modal fusion, which occurs due to the nature of depth estimation-based image projection. Yin \etal\cite{yin2024isfusion} propose two fusion approaches: hierarchical multi-modal fusion in a BEV space and instance information fusion. However, these methods still utilize a BEV space for their multi-modal fusion, which may result in a loss of 3D geometric information.

Contrary to previous approaches, Li \etal\cite{li2022unifying} recently introduce a new approach for 3D object detection that utilizes 3D voxel features. Nevertheless, this method uses dense 3D features directly without mitigating the large computational load, caused by the 3D resolution. Distinct from all the previous approaches, to the best of our knowledge, we are the first to adopt sparse 3D voxel features for 3D object detection. Our approach capitalizes on 3D geometric information while maintaining a low computational cost.
Wang \etal\cite{wang2023unitr} introduce a unified backbone for embedding multi-modal data using a transformer encoder architecture. However, they ultimately employ BEV pooling for the output features of the backbones, which may result in the loss of 3D geometric information in multi-modal features.
A recent method, CMT~\cite{yan2023cross}, utilizes BEV features for a LiDAR modality but does not explicitly transform the image features into a BEV space. Instead, they introduce a cross-modal transformer that receives image and LiDAR features as transformer key \& value pairs, and implicitly fuses them using cross-attention. However, LiDAR features are still extracted in the BEV space.

\vspace{-6pt}
\paragraph{Feature sparsification in Transformer-based detection}
DETR-based architecture has emerged in 2D object detection~\cite{detr,zhu2021deform,li2022dn,zhang2023dino,roh2022sparse,zheng2023more} and has achieved state-of-the-art performance. This architecture can be readily extended to a 3D object detector~\cite{detr3d,bai2022transfusion,chen2023futr3d,yan2023cross}. In the 2D object detection literature, several studies~\cite{roh2022sparse,zheng2023more} have aimed to reduce the number of input transformer tokens to enhance efficiency, which appears to be similar with our approach.
However, these approaches reduce the number of input features by utilizing objectness. On the contrary, our method leverages the sparse nature of the LiDAR point cloud, which does not require additional knowledge to discriminate unwanted features. To the best of our knowledge, this is the first work that employs sparse 3D voxel features directly for 3D object detection.
Furthermore, unlike the cases of 2D object detection, where detection accuracy is slightly compromised, our approach even enhances detection accuracy, demonstrating the effectiveness of utilizing sparse multi-modal features in this context.
% Furthermore, inspired from these methods, we propose an additional feature sparsification approach that eliminates redundant features, such as those corresponding to the background. This strategy can address the specific issue in our setting of inconsistent computational demands, which are caused by the varying number of transformer tokens due to the differing sparsity levels in the input data.

\section{Architecture of SparseVoxFormer}
\label{sec:baseline}

As our work present a new paradigm of 3D object detection architecture (Fig.~\ref{fig:architecture}), which directly utilizes sparse voxel features instead of BEV features, for comprehensive understanding, we first present our basic architecture essential for handling sparse features and then describe more sparse features-specific architecture. Before delving into the details, we first visit the key difference between obtaining BEV features and our sparse voxel features.
% \son{I think we need a figure of the overall framework. And after drawing this, I will modify other figures.}

% \vspace{-9pt}
\paragraph{Previous BEV-based approaches}
% BEV features (CMT~\cite{})가 어떻게 Transformer로 들어가는지 설명
As a LiDAR data for autonomous driving usually cover a wide area such as the range of [-54m, +54m] in x-, y-axes and the range of [-5m, +3m] in z-axis, raw LiDAR features need 3D voxels with a high resolution (\eg $1440\times1440\times40$) to effectively capture fine-level geometric details. However, utilizing such voxel features directly would be very cost-intensive, so that previous works reduce the feature resolution by transforming the voxel features into BEV features with the resolution of $180\times180$ (Fig. \ref{fig:key_idea}b). In this process, the voxel features can lose fine-level structural information and z-axis height information.
This LiDAR backbone is described as:
\begin{equation}
\begin{split}
    \phi_{lidar}^{bev}(L)=FCN(Sparse\_encoder_{b}(\\Sparse\_encoder_{f}(Voxelize(L)))),
\end{split}
\label{eq:bev_lidar_backbone}
\end{equation}
where \(L\) is input LiDAR sweeps and \(\phi_{lidar}^{bev}\) denotes a LiDAR backbone model of the previous BEV-based approach~\cite{yan2023cross}. \(FCN\) denotes a fully convolutional network used in \cite{yan2018second}.
For the next derivation, we divide \(Sparse\_encoder\) into front and back parts and denote them with subscripts \(f\) and \(b\), respectively.
While the backbone model contains more sub-modules, in this backbone equation, for notational simplicity, we represent key modules. The entire process will be visualized in Fig. \ref{fig:lidar_backbone}. 

% \vspace{-9pt}
\paragraph{Our approach}
% 위와 비교하여 우리는 sparse tokens을 transformer의 key and value로 곧 바로 활용할 수 있는 점을 직관적으로 설명. 
% - Transformer는 3D든 2D든 1D serialization을 하고 positional embedding을 통해 위치 정보를 부여하는 방식이기 때문에 regular topology 필요 없다. 따라서 sparse 3D data도 기존과 똑같은 방식으로 처리할 수 있다는 점을 잘 설명. 따라서 CMT에서 했던 것과 같이 기존의 DETR-like transformer decoder 구조를 활용한다.
% - Feature는 LiDAR backbone에서 BEV feature processing을 하지 않고 중간에 나온 voxel features를 활용한다.
% - Positional embedding을 voxel의 position을 통해 수행한다.

Distinct from the previous approaches that use BEV features, we directly feed sparse 3D voxel features into our 3D object detector (Fig. \ref{fig:key_idea}a).
We can easily obtain 3D voxel features with a higher resolution by omitting computations for sparse encoding (\(Sparse\_encoder_{b}\)) and BEV feature refinement (\(FCN\)) from Eq. \ref{eq:bev_lidar_backbone}:
\begin{equation}
\begin{aligned}
    &F_{lidar}=\\
    &\phi_{lidar}^{sparse}(L)=Sparse\_encoder_{f}(Voxelize(L)).
\end{aligned}
\end{equation}

The extraction of sparse 3D voxel features is straightforward. The voxel features of a sparse LiDAR point cloud are also sparse, signifying that the sparse features can be obtained by omitting zero-filled features and serializing valid feature cells as follows:
% \begin{equation}
% \begin{aligned}
    \[F_{lidar}^{sparse}=Flatten(\{f \in F_{lidar} | f \neq 0\}). \,\]
% \end{aligned}
% \label{eq:sparse_token}
% \end{equation}

\subsection{Transforemr Tokens from Sparse Voxel Features}
\label{ssec:sparse_voxel}
% \son{Should I change the title of this subsection or add some equations and details for transformer decoder?...}
% \son{And is it fine by writing description only with some figures without equations?...}

Thanks to the property of transformers that accepts serialized tokens from input data in any form, the transformer-based decoder can directly process our sparse 3D features without the need for a regular topology. \ie, we can intactly employ a previous DETR-like transformer architecture~\cite{detr} used in CMT~\cite{yan2023cross}.
We feed the sparse feature with positional embedding to the transformer decoder of 3D object detector as:
% \son{will add an equation}
\begin{equation}
\begin{aligned}
    \hat{o} =\phi_{decoder}(F'^{sparse}_{lidar}, q), 
\end{aligned}
\end{equation}
where \(F'\) denotes features \(F\) that is combined with the positional embedding, and \(\hat{o}\) and \(q\) denotes output cuboids and initial transformer queries, respectively.

Unlike CMT, the positional part for keys can be directly encoded into 3D positional embedding \( E_{pos}(x, y, z) \) by using the voxel feature coordinates \( (x, y, z) \). Then, the combined feature \(F'\) is defined by:

\[ F' = F + E_{pos}(x, y, z). \]

The query \(q\) for the first transformer decoder layer is in the form of a learnable vector following DETR and CMT, constructed by the same positional embedding \(E_{pos}\) of randomly initialized of 3D coordinates of \( (x, y, z) \). These coordinates are trained in the training phase and fixed in the testing phase.
In this transformer-based architecture, the computational complexity is almost proportional to the number of key \& value tokens. Therefore, reducing the number of tokens by receiving sparse features consequently reduces the computational cost.

% - Attn order change: 먼저 query enhancement 전에 attn order에 대해 논의하겠다. 현재 구조로는 query가 refinement가 되어 좋아지더라도 cross-attn 전에 self-attn이 들어가기 때문에 refined query가 self-attn을 통해서 변경된다. 따라서 우리가 의도한 cross-attn이 불가하다. Mask2Former, DINO 등을 보면 일반적인 구조와 다르게 cross-attn 후에 self-attn을 사용한다. 앞선 work에서는 이러한 구조에 대한 설명이 없었다. Query가 충분히 좋다고 가정했을 때 cross-attn을 수행하고 cross-attn의 결과를 self-attn을 통해 refine하는 게 practically 좋다고 생각된다.

% - Query initialization via core-set selection: scene-agnostic 고정된 query 쓸 필요 없이, sparse tokens으로 부터 scene-adaptive dynamic initial query를 만들 수 있음. 그러나 LiDAR point는 멀수록 sparse하기 때문에 단순히 random sampling을 하게 되면 원거리 정보들이 무시될 수 있다. 이러한 문제를 해결할 수 있도록 정보의 rare density를 효과적으로 고려하면서 sampling할 수 있는, active learning에서 사용되던 core-set selection appraoch를 query selection에 도입했다.

% - query refinement for precise attention: to exploit rich geometric features in sparse voxels, a query should have a precise reference position. However, CMT, which follows on the basic DETR approach, use fixed learnable query.
% Inspired by Deformalbe DETR and DAB-DETR, we adopt query refinement approach.
% This enables cross-attn exploiting rich geometric features in key \& value along with more precise positional embedding.

%%%--------------------------------------------------------------------------------
\begin{figure}[t]
\begin{center}
\includegraphics [width=0.8\linewidth] {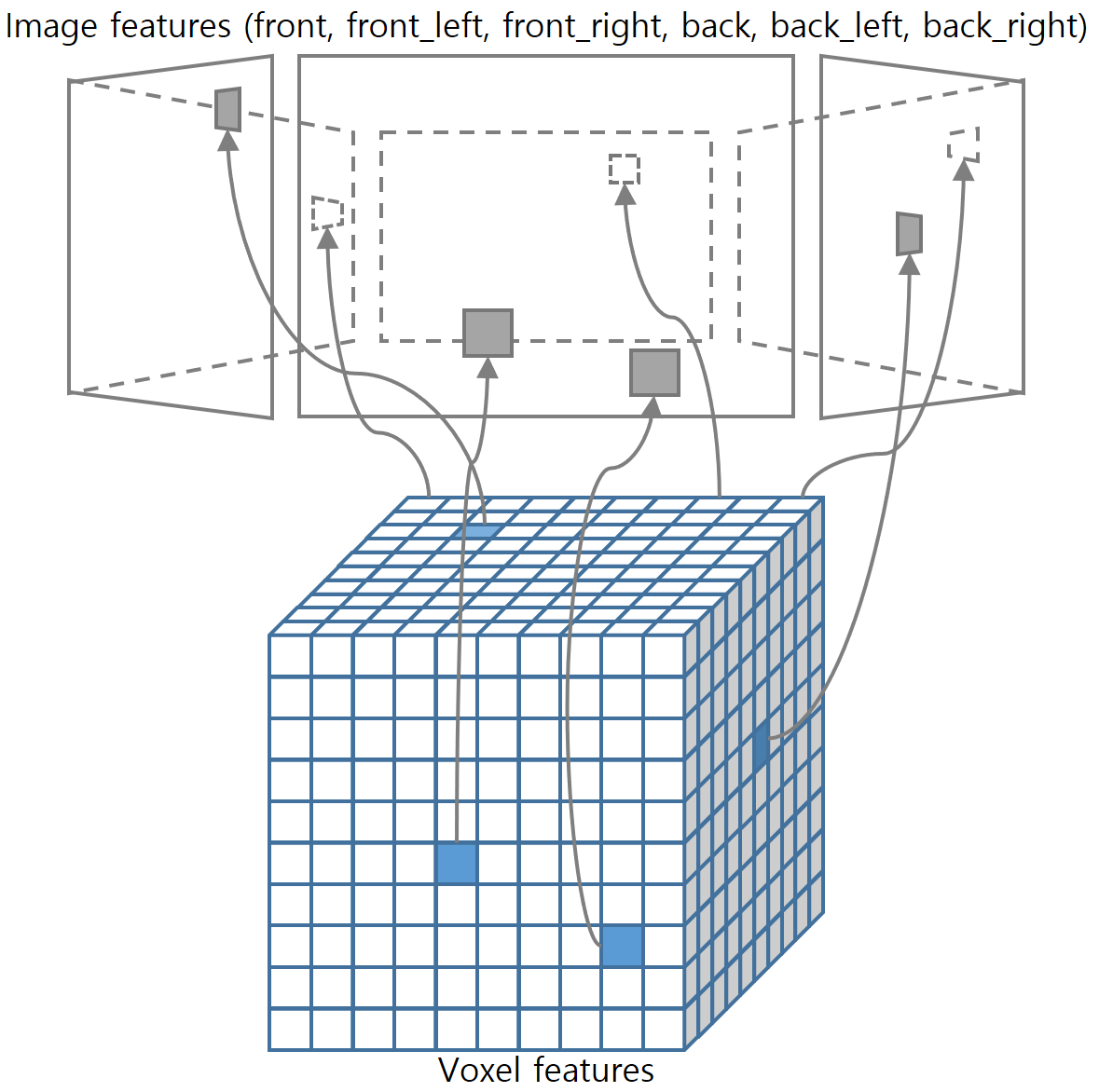}
\end{center}
\vspace{-10pt}
  \caption{Explicit multi-modal fusion without image depths is available in our voxel-based approach since each valid cell in 3D voxels already possess 3D coordinates, required for LiDAR to camera transformation. A LiDAR point can be easily projected to the camera space by a pre-defined LiDAR-camera transformation matrix. Similarly, each valid voxel feature has a corresponding projected image feature by the same transformation matrix.
}
\label{fig:explicit_fusion}
\vspace{-4pt}
\end{figure}
%%%--------------------------------------------------------------------------------

%%%--------------------------------------------------------------------------------
\begin{figure*}[t]
\begin{center}
\includegraphics [width=0.90\linewidth] {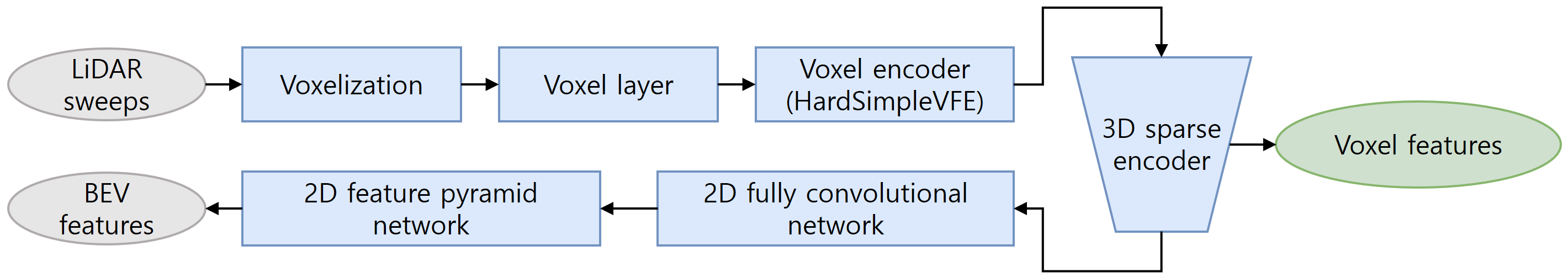}
\end{center}
\vspace{-10pt}
  \caption{Feature processing to produce BEV features in a LiDAR backbone in CMT~\cite{yan2023cross}. Our sparse voxel features are intermediate results in this process (in 3D sparse encoder). Voxelization, voxel layer, and voxel encoder do not contain learnable parameters. 
  % The output of sparse encoder is a dense feature map. 
  % \son{should be modified.}
  % Note that larger voxel resolution uses less computations.
}
\label{fig:lidar_backbone}
\vspace{-8pt}
\end{figure*}
%%%--------------------------------------------------------------------------------

% \subsection{Explicit multi-modal fusion with sparse voxel features}
\subsection{Explicit Multi-modal Fusion with Sparse Features}
\label{ssec:explicit_fusion}
% 각 cell은 3D coordinates를 가지고 있기 때문에 정확하게 image feature를 가져와서 결합 가능
In our sparse voxel-based approach, multi-modal fusion with image features is more intuitive than in BEV-based approaches. 
For instance, Liang \etal\cite{liang2018} handle the one-to-many mapping between a BEV point and image points by adopting the concept of continuous fusion.
In contrast, since our 3D voxel features carry their 3D positional coordinates \( (x, y, z) \), they can be accurately projected to image feature space by:
\[ (u, v, 1)^T = K \cdot T_{LiDAR \rightarrow Camera} \cdot (x, y, z, 1)^T, \]
where \( T_{LiDAR \rightarrow Camera} \) and \( K \) denote \(4\times4\) LiDAR-to-camera transform matrix and camera intrinsic parameters, respectively. This process is depicted in Fig. \ref{fig:explicit_fusion}. 
% It is well-known that LiDAR points can be easily projected to an image and play a role of image depths. Our explicit fusion uses this nature.

This implies that we can obtain paired features of LiDAR and image features (\(F_{lidar}^{sparse}\) and \(F_{image}^{(u, v)}\), respectively) while preserving the sparsity by:
% One can argue that
% 실험적으로 LiDAR가 생각보다 원거리 잘가져옴
% 적게 찍혀서 false positive가 늘어나는거가 문제점임
% 이러한 ambuiguity를 image modality가 보강해주는데, 사실 lidar가 찍힌 cell에만 image feature를 제공해줘도 empirically 물체 인식에 큰 문제 없음
\begin{equation}
\begin{aligned}
    F^{sparse}_{combined} = Concat(F_{lidar}^{sparse}, F_{image}^{(u, v)}).
\end{aligned}
\label{eq:explicit_fusion}
\end{equation}

We found that fusing these multi-modal features by only a simple concatenation significantly enhance detection performance. It does not increase a computational overhead in the transformer decoder because it does not change the number of input tokens while CMT does increase.

\subsection{Multi-modal Sparse Feature Refinement}
\label{ssec:dsvt}
% \vspace{-3pt}
\paragraph{Necessity of additional feature refinement}
We found that our na\"ive LiDAR backbone would be insufficient to fully encode fine-level geometric features as we only use partial computations (about 30\%) of a LiDAR backbone, as shown in Fig.~\ref{fig:lidar_backbone} and will be analyzed in Table~\ref{tbl:comp_cost}.
The computations are primarily used for adjusting the 3D resolution of voxel features in the sparse encoder, and computations for feature refinement are absent.
% \son{need to revise.}
% Specifically, we use a minimum of 3D convolution layers to transform a point cloud to a voxel and omit a BEV feature processing containing the majority of 3D/2D convolution layers in the LiDAR backbone.
% While we supplement this by presenting the modified voxel feature encoding (mVFE), our new backbone is still too na\"ive to use only about 30\% of parameters of the conventional LiDAR backbone for BEV features although our feature rather has a much larger resolution, as will be analyzed in Table~\ref{tbl:comp_cost}.
To remedy this, we employ DSVT~\cite{wang2023dsvt} for our voxel feature refinement, which can refine sparse features while preserving the sparsity of the features. After passing DSVT blocks, the result LiDAR features have richer geometric information. 
% We found that this model fully exploit the benefit of the increase of voxel resolution (Sec. \ref{ssec:voxel_resolution}).

% \vspace{-7pt}
\paragraph{Deep fusion module}
In our multi-modal fusion approach, we explicitly combine LiDAR and camera features while preserving the sparsity of the LiDAR features. In this case, we can also consider applying DSVT not only to the LiDAR features but to the fused features because the fused features now have the same form of the LiDAR features. By doing so, the additional sparse feature refinement module can facilitate the multi-modal fusion as well as refine sparse features for encoding fine-level geometric information.

To this end, we introduce a deep fusion module (DFM) to apply the DSVT module into the multi-modal fused features. 
Previous methods typically employ single modality-wise feature refinement, followed by multi-modal fusion accomplished through concatenation of the refined features.
On the other hand, we directly refine our multi-modal sparse features by extending Eq.~\ref{eq:explicit_fusion} to:
\begin{equation}
    F^{sparse}_{combined} = DFM(Concat(F^{sparse}_{lidar}, F_{image}^{(u, v)})),
\end{equation}
where \(DFM\) denotes the deep fusion module, which consists of a sequence of DSVT blocks~\cite{wang2023dsvt}.
\begin{table*}[t]
\centering
% \small
\scalebox{0.935}{
\begin{tabular}{ c | c c | c | c | c | c}
\toprule
Model & ~mAP~ & ~NDS~ & Feat. Res &\# of tokens & Detector cost (GFLOPs) & Model size (M) \\
\toprule
UVTR~\cite{li2022unifying} & 65.4 & 70.2 & $180\times180\times5$ & 162,000 & - & 88.9 \\
CMT~\cite{yan2023cross} & 70.3 & 72.9 & $180\times180\times1$& 62,400 & 163.6 & 83.9 \\
% Our baseline w/o mVFE &  70.7 & 72.9 & $180\times180\times11$ & 18,000 & 77.5 \\
SparseVoxFormer-base  & \textbf{70.8} & \textbf{73.2} & $180\times180\times\textbf{11}$ & \textbf{18,000} & \textbf{61.3} &\textbf{77.5} \\
\bottomrule
\end{tabular}
}
\vspace{-8pt}
\caption{Effect of sparse multi-modal features on the nuScenes val set~\cite{caesar2020nuscenes}.}
\vspace{-10pt}
\label{tbl:baseline}
\end{table*}
%%%--------------------------------------------------------------------------------

%%%--------------------------------------------------------------------------------
\begin{table}[t]
\centering
\small
\scalebox{0.74}{
\begin{tabular}{ c | c c | c c }
\toprule
\multirow{2}{*}{Application of DSVT}& \multicolumn{2}{c|}{CMT} & \multicolumn{2}{c}{SparseVoxFormer}  \\
& mAP & NDS & mAP & NDS\\
\toprule
None & 70.3 & 72.9 & 70.8 & 73.2  \\
LiDAR & 70.8\textcolor{red}{($+0.5$)}  & 73.2\textcolor{red}{($+0.3$)} & 71.7\textcolor{red}{($+0.9$)} & 74.4\textcolor{red}{($+1.2$)} \\
Multi-modal (deep fusion) & - & - & \textbf{72.3}\textcolor{red}{($+1.5$)} &  \textbf{74.5}\textcolor{red}{($+1.3$)} \\
\bottomrule

\end{tabular}
}
\vspace{-8pt}
\caption{Effect of sparse feature refinement via DSVT (deep fusion) according to an applied modality. 
% \son{(without \textbf{feature elimination})}
% \son{Can we replace all models with our small version?}
}
\vspace{-8pt}
\label{tbl:dsvt}
\end{table}
%%%--------------------------------------------------------------------------------

\subsection{Redundant Feature Elimination}
\label{ssec:feature_elimination}
% \son{We should refer to Sparse-DETR, Focus-DETR.}
% \textbf{rewrite}
Despite our basic voxel sparsification which removes more than 90\% features in the original voxel features, the number of fully sparse features may still introduce redundant computations, particularly since many of these features pertain in backgrounds like buildings and roads, which are irrelevant for 3D object detection.
Moreover, distinct from the 2D object detection case~\cite{roh2022sparse,zheng2023more}, the different sparsity of LiDAR data may cause computational instability by producing different number of transformer tokens.
To handle these problems, inspired by \cite{roh2022sparse,zheng2023more}, we present an additional feature elimination scheme which removes the majority of our sparse features before they are fed into the detector. %that contributes marginally to the query attention update.

To this end, we employ an auxiliary binary classification head before the transformer decoder. 
Profiting from the nature of our sparse features, each of which carries central coordinates $(x,y,z)$, each feature can be directly supervised by whether the coordinates is belong to the cuboids of object detection annotations.
To train the auxiliary head, we utilize focal loss~\cite{8417976} with a binary label. A label is assigned a value of 1 if a voxel feature belongs to any bounding cuboid of the annotations, and vice versa.
To prevent a true negative case, we use more generous positive labels by dilating the size of the cuboids by $50\%$.
We can eliminate redundant background features by retaining the Top-K features based on the confidence score of the trained head, implying that the detector uses a fixed number of tokens.

\section{Experimental Results}
\label{sec:experiment}
% \vspace{-5pt}
\paragraph{Implementation details}
Following to CMT, we use bipartite matching~\cite{detr} for set prediction, focal loss~\cite{8417976} for classification , L1 loss for cuboid regression, and query denoising~\cite{li2022dn}.
We use the nuScenes training dataset~\cite{caesar2020nuscenes} and train our models for 20 epochs, specifically with GT sampling for first 15 epochs and without the sampling for later 5 epochs.
% With 8 NVIDIA A100 GPUs, the training spends about ...
We use VoVNet~\cite{lee2019energy} as a image backbone, and a part of VoxelNet~\cite{DBLP:conf/cvpr/ZhouT18} as a LiDAR backbone as shown in Fig. \ref{fig:lidar_backbone}.
% \supp{Training and architectural details can be found in the supplementary material.}
% Regarding the Transformer decoder, we use 6 transformer decoder layers.
% Regarding a deep fusion module, we use four DSVT blocks~\cite{}.
We use voxel features with the final resolution of $180\times180\times11$ with the input voxel size of 0.075m for the following experiments unless we notify.

%%%--------------------------------------------------------------------------------
\begin{table}[t]
% \centering
\scalebox{0.75}{
\begin{tabular}{ c | c c }
\toprule
~~\# of tokens~~ & ~~~mAP~~~ & ~~~NDS~~~\\
\toprule
Full & 72.3 & 74.5 \\
Half & 72.2 & 74.4 \\
10,000 & 72.2 & 74.4 \\
7,500 & 72.1 & 74.3 \\
5,000 & 70.0 & 73.4\\
2,500 & 55.5 & 65.8 \\
\bottomrule
\end{tabular}
}
\scalebox{0.75}{
\begin{tabular}{c | c c}
\toprule
Voxel size (m) & ~mAP~ & ~NDS~  \\
\toprule
% CMT~\cite{yan2023cross} & 0.075 & 70.3 & 72.9\\
% DAL~\cite{huang2023detecting} & 0.05 &  71.5 & 74.0\\
% \midrule
0.20 & 70.5 & 72.7 \\
0.10 & 71.3 & 73.7 \\
0.075 & 72.2  & 74.4   \\
0.05 & \textbf{72.5} & \textbf{74.9}  \\
\bottomrule
\end{tabular}
}
\vspace{-8pt}
\caption{Accuracy according to the number of remaining transformer tokens (left) and input voxel resolutions (right).}
\label{tbl:sparsification}
\vspace{-6pt}
\end{table}
%%%--------------------------------------------------------------------------------

%%%--------------------------------------------------------------------------------
\begin{table*}[t]
\centering
\small
\scalebox{0.95}{
\begin{tabular}{ c | c c | c c | c c c c}
\toprule
\multirow{2}{*}{}& \multicolumn{2}{c|}{LiDAR backbone}  & \multicolumn{2}{c|}{Decoder (Detector)} & \multicolumn{3}{c}{Total}\\
& params & cost & params & cost & mAP & NDS & cost$^*$ \\ %& latency \\
\toprule
CMT~\cite{yan2023cross} & 8.5 & 155.1 & 4.8 & 163.6 & 70.3 & 72.9 & 318.7 \\ %& 179.7  \\
SparseVoxFormer-base & 2.5\textcolor{red}{($-71\%$)} & 50.0\textcolor{red}{($-68\%$)} & 4.5 & 61.3\textcolor{red}{($-63\%$)} & \textbf{70.8} & \textbf{73.2} & \textbf{111.3}\textcolor{red}{($-65\%$)} \\ %& \textbf{147.9}  \\
\midrule
% Ours-S & 70.6 & 2043.9 & 4.6\textcolor{red}{($-46\%$)} & 93.0\textcolor{red}{($-40\%$)} & 4.6 & 39.9\textcolor{red}{($-76\%$)} & 71.4 & 73.4 & 160.5\\
CMT w/ DSVT & 18.0 & 347.5 & 4.8 & 163.6 & 70.8 & 73.2 & 511.1 \\ %& 212.8 \\
SparseVoxFormer & 12.1\textcolor{red}{($-33\%$)} & 242.4\textcolor{red}{($-30\%$)} & 4.6 & 39.9\textcolor{red}{($-76\%$)}& \textbf{72.2}& \textbf{74.4} & \textbf{282.3}\textcolor{red}{($-45\%$)} \\ % & \textbf{179.2}  \\
% Ours (final, 0005) &  &  &  & & & & & & -  \\
\bottomrule

\end{tabular}
}
\vspace{-8pt}
\caption{Module-wise computational cost analysis. Params and cost means the number of parameters (M) and the computational cost (GFLOPs), respectively. Since the camera backbone part (VoVNet~\cite{lee2019energy}) is identical across all variants, the details are omitted. $^*$For the total cost, the common cost for camera backbone is excluded for clearer comparison.
}
\vspace{-8pt}
\label{tbl:comp_cost}
\end{table*}
%%%--------------------------------------------------------------------------------

% \vspace{-9pt}
\paragraph{Evaluation metrics}
We use two evaluation metric in this paper. First is mAP (mean Average Precision), which is similar to 2D object detection, but defined by using an overlap between 3D cuboids of a prediction and its label instead 2D boxes of them.
Second is NDS (Nuscenes Detection Score), which considers five factors: translation, scale, orientation, velocity, and attribute errors of the cuboid of each instance.

\subsection{Component Analysis}
% \vspace{-3pt}
\paragraph{Effectiveness of using sparse voxels (baseline)}
This paper presents a new paradigm to use sparse 3D voxel features from the multi-modal input of LiDAR and cameras for 3D object detection.
Our SparseVoxFormer uses high-resolution features more effectively than UVTR~\cite{li2022unifying}, which utilizes full 3D voxel features, thanks to the sparse nature of LiDAR data (refer to Table~\ref{tbl:baseline}).

Furthermore, compared to the state-of-the-art BEV-based model (CMT~\cite{yan2023cross}), the only modifications from the CMT model to adopt the use of multi-modal sparse features allow our base model to achieve higher accuracy (mAP) while significantly reducing the number of transformer tokens. These experimental results demonstrate that directly using sparse voxel features with a higher resolution is not only a feasible approach for efficiently handling 3D object detection, but it can also be a more effective method than previous BEV feature-based approaches.
%While the essential components have already shown sufficient improvement over previous BEV-based paradigm

% \vspace{-9pt}
\paragraph{Effect of sparse feature refinement}
% \son{Redesign of the experiment section may be required. This experiment is required, but also overlapped with the main ablation study.}
In this section, we show how the additional sparse feature refinement effectively improves the detection accuracy of our models. Although our approach uses fewer computations for the LiDAR backbone compared to previous methods, one might still question if the benefits of using additional computations of DSVT can be applied to other approaches. To address this question, we compare CMT models with our model variants.
% experiment detail
% When we apply DSVT modules to our model, we increase the target number of channels of our sparse encoder twice (128 to 256) because CMT also use 256 channels for BEV feature refinement as well as most recent models such as DAL~\cite{huang2023detecting} also use 256 channels for sparse encoding. We empirically found that 256 channels is minimally required for effective feature encoding.

% desired result
% In the case of CMT, CNNs for BEV feature extraction has already processed sufficient geometric information so that the gain by employing the additional DSVT is relatively smaller than that of our model (Table \ref{tbl:dsvt}).
% On the other hand, as we use a minor part of the previous LiDAR backbone used in CMT, our sparse features may not be sufficiently encoded. Therefore, sparse feature refinement via DSVT brings significant performance improvement.
% It is noteworthy that the margin versus CMT becomes further larger by our deep fusion approach exploiting the sparse nature of our multi-modal fused features, realized by our explicit multi-modal fusion.
% It is surpriging that the deep fusion module simply relocate the DSVT blocks without any computational overhead.
% This implies that refinement of sparse features encodes rich geometric information as well as facilitates the multi-modal fusion.
In the case of CMT, the CNNs for BEV feature refinement have already processed sufficient geometric information, resulting in a relatively smaller gain when employing the additional DSVT compared to our model (Table \ref{tbl:dsvt}). On the other hand, as we utilize only a minor part of the previous LiDAR backbone used in CMT, our sparse features may not be sufficiently encoded. Therefore, sparse feature refinement via DSVT brings significant performance improvement. It is noteworthy that the performance gap compared to CMT becomes even larger thanks to our deep fusion approach that exploits the sparse nature of our multi-modal fused features, realized by our explicit multi-modal fusion. Interestingly, the deep fusion module simply relocates the DSVT blocks without any computational overhead, implying that our refinement of sparse features not only encodes rich geometric information but also facilitates the multi-modal fusion.
% In addition, thanks to our explicit multi-modal fusion, we can apply DSVT to the multi-modal fused features directly, and this effectively facilitates the multi-modal fusion and further enhance the performance. 

% \vspace{-9pt}
\paragraph{Effect of feature elimination}

Our feature elimination scheme can effectively reduce the number of transformer tokens (from 18,000 in average) while almost preserving the original performance (Table \ref{tbl:sparsification} left). 
Specifically, the performance almost never drops when we use 50\% of tokens or the fixed number of 10,000 tokens, and we use 10,000 tokens in later experiments.
% Specifically, the performance in mAP and NDS changes from (72.3, 74.5) without feature elimination to (72.2, 74.4) with feature elimination.
We want to note that the overhead of an auxiliary head for the feature elimination is marginal.
It is noteworthy that the use of the fixed number of tokens enhances the practical usefulness of our approach by regularizing a consistent computational cost.
% Fig.~\ref{fig:intro}b shows that our feature elimination properly removes redundant features on backgrounds such as roads.

% \vspace{-9pt}
\paragraph{Input voxel resolutions}
\label{ssec:voxel_resolution}
% \son{This experiment would be better to be placed in the main paper if a paper limit allows.}
According to input voxel resolutions, our model performance can be further improved (Table \ref{tbl:sparsification} right).
A smaller voxel size means a larger input voxel resolution for a LiDAR modality.
% Table \ref{tbl:voxel_size} shows that our model with the voxel size of 0.075m shows a higher accuracy than previous state-of-the-art models with a smaller voxel size. The margin becomes larger with our model with the smaller size (0.05m).
% Table \ref{tbl:sparsification} (right) shows that our models with larger voxel resolutions achieve higher performance.
% \supp{Deeper analysis on additional model variants such as according to different backbone and resolution for a camera modality can be found in the supplementary material.}

% \paragraph{Ablation study}
% \son{can consider to remove ablation study and add Effect of Feature Elimination}
% On top of the baseline architecture, we propose advanced modules to further improve our architecture exploiting sparse voxel features.
% In this section

%%%--------------------------------------------------------------------------------
% \jia{I list all experiments results and other SOTA method}
\begin{table*}[t]
\centering
\small
\scalebox{0.9}{
\begin{tabular}{ l | c | c | c c c c | c}
\toprule
Methods & Present at & Modality & mAP(val) & NDS(val) & mAP(test) & NDS(test) & latency (ms) \\
\toprule
PointPainting~\cite{vora2020pointpainting} & CVPR'20 & C+L & 65.8 & 69.6 & - & - & -\\
% PointAugmenting~\cite{wang2021pointaugmenting} & CVPR'21 & C+L & - & - & $66.8^\dag$ & $71.0^\dag$ & -\\
MVP~\cite{yin2021multimodal} & NeurIPS'21 & C+L & 66.1 & 70.0 & 66.4 & 70.5 & -\\
TransFusion~\cite{bai2022transfusion} & CVPR'22 & C+L & 67.5 & 71.3 & 68.9 & 71.6 & - \\
AutoAlignV2~\cite{chen2022autoalignv2} & ECCV'22 & C+L & 67.1 & 71.2 & 68.4 & 72.4 & -\\
UVTR~\cite{li2022unifying} & NeurIPS'22 & C+L & 65.4 & 70.2 & 67.1 & 71.1 & 264\\
BEVFusion(PKU)~\cite{liang2022bevfusion} & NeurIPS'22 & C+L & 67.9 & 71.0 & 69.2 & 71.8 & -\\
DeepInteraction~\cite{yang2022deepinteraction} & NeurIPS'22 & C+L & 69.9 & 72.6 & 70.8 & 73.4 & 594\\
FUTR3D~\cite{chen2023futr3d} & CVPRW'23 & C+L & 64.5 & 68.3 & - & -  & -\\
BEVFusion(MIT)~\cite{liu2023bevfusion} & ICRA'23 & C+L & 68.5 & 71.4 & 70.2 & 72.9 & 221\\
CMT~\cite{yan2023cross} & ICCV'23 & C+L & 70.3 & 72.9 & 72.0 & 74.0& 180\\ 
UniTR~\cite{wang2023unitr} & ICCV'23 & C+L & 70.5 & 73.3 & 70.9 & 74.5 & 196 \\
ECFusion~\cite{fu2024} & ICRA'24 & C+L & 70.7 & 73.4 & 71.5 & 73.9 & - \\ 
ISFusion~\cite{yin2024isfusion} & CVPR'24 & C+L & \textbf{72.8} & 74.0 & \textbf{73.0} & 75.2 & 214\\
GraphBEV~\cite{song2024graphbev} & ECCV'24 & C+L & 70.1 & 72.9 & 71.7 & 73.6 & 234\\
\midrule
SparseVoxFormer & - & C+L & 72.2 & \textbf{74.4} & 72.9 & \textbf{75.3} & \textbf{179} \\
\bottomrule
\end{tabular}
}
\vspace{-8pt}
\caption{Performance comparison in 3D object detection on nuScenes (val and test sets)~\cite{caesar2020nuscenes}.
% We denote the best values in red, and the second best values in blue.
Latency (ms) is measured by averaging the inference times over first 1,000 samples of nuScenes val set using a single NVIDIA A100 GPU. Notion of modality: Camera (C), LiDAR(L).
}
\vspace{-10pt}
\label{tbl:comparison}
\end{table*}
%%%--------------------------------------------------------------------------------

%%%--------------------------------------------------------------------------------
\begin{table}[t]
\centering
\small
\scalebox{0.74}{
\begin{tabular}{ c | c c | c c | c c | c c }
\toprule
\multirow{2}{*}{}& \multicolumn{2}{c|}{\footnotesize{SemanticBEVFusion~\cite{jiang2023semanticbevfusion}}}& \multicolumn{2}{c|}{CMT} & \multicolumn{2}{c|}{~Ours-base~}  & \multicolumn{2}{c}{Ours} \\
& ~~~mAP~~~ & ~~~NDS~~~ & mAP & NDS & mAP & NDS & mAP & NDS \\
\toprule
Whole & 69.5 & 72.0 & 70.3 & 72.9 & 70.8 & 73.2 & \textbf{72.2} & \textbf{74.4} \\
Near & 79.9 & 78.1 & 80.5 & 78.9 & 80.3 & 79.1  & \textbf{82.9} & \textbf{81.0} \\
Middle& 66.1 & 70.2 & 66.5 & 71.0 & 67.3 & 71.1 & \textbf{68.3} & \textbf{72.0} \\
Far & 37.0 & 49.1 & 37.3 & 49.3  & 38.2 & 49.9  & \textbf{39.4} &\textbf{50.9}\\
\bottomrule

\end{tabular}
}
\vspace{-8pt}
\caption{Range-wise performance comparison.
% \son{I will complete this table in the next week.}
}
\vspace{-6pt}
\label{tbl:range_performance}
\end{table}
%%%--------------------------------------------------------------------------------

%%%--------------------------------------------------------------------------------
\begin{table}[t]
\centering
% \small
\scalebox{0.8}{
\begin{tabular}{ c | c | c c }
\toprule
Img. backbone & Model & ~~mAP~~ & ~~NDS~~ \\
\toprule
% \multirow{3}{*}{ None (LiDAR only)} & VoxelNeXt~\cite{chen2023voxenext} &56.2 & 64.3  \\
\multirow{3}{*}{\shortstack{None\\(LiDAR only)}} & VoxelNeXt~\cite{chen2023voxenext} &60.5 & 66.6  \\
 & CMT~\cite{yan2023cross} &62.1 & 68.6  \\
 & SparseVoxFormer &\textbf{65.7} & \textbf{70.9}\\
% \midrule
% ResNet-18 & SparseVoxFormer & & \\
\midrule
\multirow{2}{*}{ ResNet-50} & CMT~\cite{yan2023cross}  & 67.9 & 70.8 \\
 & SparseVoxFormer  & \textbf{69.3} & \textbf{72.7} \\
\midrule
\multirow{2}{*}{VoVNet}  & CMT~\cite{yan2023cross} & 70.3 & 72.9\\
  & SparseVoxFormer & \textbf{72.2} & \textbf{74.4} \\
\bottomrule
\end{tabular}
}
\vspace{-8pt}
\caption{Performance evaluation according to different image backbones on the nuScenes val set~\cite{caesar2020nuscenes}.}
\vspace{-8pt}
\label{tbl:image_backbone}
\end{table}
%%%--------------------------------------------------------------------------------

%- Additional components: 1) Query selection from sparse voxel tokens, 2) Query refinement, 3) Modified voxel encoding.
\subsection{Computational Cost Analysis}
% LiDAR backbone, image backbone, transformer 등 module 별 compuational cost 분석
An important benefit of our approach is its ability to exploit rich geometric information in high-resolution 3D features while consuming a relatively small computational cost. Consequently, the computational cost analysis would be essential to show the effectiveness of our architecture.
For highlighting the contrast, we conduct a module-wise comparison of our model with CMT~\cite{yan2023cross}, which, to the best of our knowledge, is one of the fastest among state-of-the-art models. CMT is also suitable for an apple-to-apple comparison with our method due to its similar module composition. We note that, for simplicity, we include the computational cost of DSVT modules for sparse feature refinement to a LiDAR backbone.

% Regarding evaluation metrics, we use the computational cost in FLOPs and the number of parameters of multi-modal 3D detection models.
% An inference speed may not fully reflect this theoretical computational cost due to different optimization levels for various operations, especially for handling sparse features.
% Therefore, we believe that evaluating in FLOPs is theoretically meaningful. 

% Note that there has been no well-established approach to calculate the computational complexity of multi-modal 3D object detector and models with sparse data and sparse operations to our best knowledge. Therefore, we separately inference LiDAR-only-based and camera-only-based branches, and manually combine their values. Furthermore, the computational cost of sparse convolution-based layers can vary according to the sparsity of LiDAR data. We assume the number of valid intermediate features as 20,000 in the layers for simplicity. For simplicity, we include the computational cost of DSVT modules to a LiDAR backbone.

Table \ref{tbl:comp_cost} shows that our architecture substantially reduces computational costs while even enhancing detection performance.
% Regrading the decoder, our approach shows a significantly smaller cost than that of CMT because our voxel features yield around 18,000 tokens in average while CMT yields 62,400 transformer tokens (32,000 for LiDAR and 24,000 for 6 cameras). Thanks to our feature elimination scheme, we can further reduce the number of transformer tokens to the fixed number of 10,000, reducing also the computational cost of our transformer-based detector.
% Regarding the LiDAR backbone, as our approach uses 3D voxel features obtained from an early stage of the backbone, the computational cost for BEV feature processing is much saved. Despite these simply encoded 3D geometric features, our base SparseVoxFormer (Ours-base) shows a better accuracy than that of CMT. With the additional sparse feature refinement, the margin between our final SparseVoxFormer (Ours-S) and CMT becomes significant larger. 
We emphasize again that merely incorporating the DSVT module into the existing model (CMT) is not as effective. The CMT model with DSVT has a cost for a LiDAR backbone and a detector that is five times larger than our base model, yet it only shows comparable performance. Our final model surpasses both CMT models and does so at a lower cost.
It is also noteworthy that the effect of this cost saving will be more effective in the embedded environment, which is limited to use smaller backbones.

\subsection{Comparison}
% -Other previous methods
% -BEVFusion, Deep Interaction, CMT
% -UVTR
% - without any test-time augmentation (TTA), model ensembles
Finally, we present a comprehensive comparison (Table~\ref{tbl:comparison}) with the state-of-the-art multi-modal 3D object detection models~\cite{vora2020pointpainting,yin2021multimodal,bai2022transfusion,chen2022autoalignv2,li2022unifying,liang2022bevfusion,yang2022deepinteraction,chen2023futr3d,liu2023bevfusion,yan2023cross,wang2023unitr,fu2024}.
% Except PointAugmenting~\cite{wang2021pointaugmenting}, all models do not use any test-time augmentation and model ensembles for this evaluation.
All models do not use any test-time augmentation and model ensembles for this evaluation.
% Our model has achieved a new state-of-the-art performance while showing the fastest speed.

Although our approach pursues an efficiency-oriented design, which focus to reduce the number of transformer tokens based on the sparsity of LiDAR data, our model shows comparable or even better performance than recent state-of-the-art methods~\cite{fu2024,yin2024isfusion} employing complicated multi-stage architectures including instance-level information processing for higher accuracies.

% As our architecture is based on the Transformer decoder, it is fully compatible with well-studied DETR-based components. For instance, when iterative query refinement (IR)~\cite{} is adopted by our model, the performance easily increases without additional cost overhead.
% Finally, our model, SparseVoxFormer w/ IR, has achieved a new state-of-the-art performance while demonstrating the fastest speed.

\paragraph{Range-wise performance}
Our explicit multi-modal fusion uses sparse image features instead of dense ones. As a result, there might be questions regarding the performance of our models concerning long-range object detection. Generally, it is understood that detection of long-range object detection depends more on the camera modality, since such objects may contain only a few LiDAR points~\cite{jiang2023semanticbevfusion}. To address these questions, we measure the range-wise accuracy of 3D object detection.
Following SemanticBEVFusion~\cite{jiang2023semanticbevfusion}, we divide the detection range of 0\(\sim\)54m from an ego vehicle into near (0\(\sim\)18m), middle (18\(\sim\)36m), and far (36\(\sim\)54m) ranges.

Our approach outperforms previous approaches in all the ranges (Table~\ref{tbl:range_performance}). It is noteworthy that, as shown in the comparison of CMT~\cite{yan2023cross} and Ours-base, our architecture enhances a long-range performance from CMT more than near- and mid-range performances.
% \son{I am working on implementing an evaluation code of range-wise performance.}

\paragraph{Smaller image backbones}
\label{ssec:smaller_backbones}
In this work, we focus on the LiDAR backbone and transformer decoder suitable for SparseVoxformer, while adhering to the settings of CMT~\cite{yan2023cross} regarding the image backbone. However, as shown in a previous experiment on computational cost analysis in Table~\ref{tbl:comp_cost}, the cost of an image backbone is significant in the case of using VoVNet~\cite{lee2019energy}. To demonstrate the practical utility of our approach in fields requiring lower computational resources, we have prepared additional model variants without an image backbone (LiDAR only) or with a smaller image backbone (ResNet-50~\cite{resnet}).
Table~\ref{tbl:image_backbone} shows that our variants of SparseVoxFormer outperform all the variants of CMT.
It is noteworthy that our LiDAR-only model still outperform the previous state-of-the-art LiDAR-based model~\cite{chen2023voxenext}.

\section{Conclusion}
\label{sec:conclusion}

This paper introduces a new architecture, referred to as SparseVoxFormer, for multi-modal 3D object detection. This is based on our key observation that using 3D voxel features with higher resolution can result in smaller computational expenses compared to using BEV features, by leveraging the sparse nature of LiDAR data. With our foundational designs for integrating sparse voxel features into a transformer-based detector and explicit multi-modal fusion with the sparse features, our SparseVoxFormer exhibits a performance comparable to previous state-of-the-art approaches, but with significantly reduced computational cost. We further facilitate more sparse but well-fused multi-modal features through a deep fusion module and an additional feature elimination scheme. Consequently, our approach achieves state-of-the-art performance on the nuScenes dataset with a faster inference speed.
% Additional discussion on limitation and future work can be found in the supplementary material.

% \input{07_supp}

%%%%%%%%% REFERENCES
{\small
\bibliographystyle{ieee_fullname}
\bibliography{egbib}
}

\clearpage
\appendix

\section{Additional Implementation Details}
\subsection{Voxel Encoding for Rich Geometric Features}
\label{ssec:mvfe}
% 우리 방법은 이론적으로 high-resolution space를 사용할 수 있지만, 현재 기존 방법들이 사용하고 있는 voxel encoding (HardSimpleVFE) 방법은 voxel내 point들을 단순히 average하기 때문에 fine geometric detail 손실되어 우리 방법의 merit을 exploit하기 어렵다.
% 이를 완화하기 위하여 average와 함께 std를 같이 계산하도록 간단하게 voxel encoding 방법을 수정했다.
Due to the nature of our approach that directly exploits geometric information in the sparse voxel features, it is important that the features encodes rich geometric information. However, our voxel features are yielded from the intermediate stage of a LiDAR backbone (Fig. 4 in the main paper) so that each voxel feature may not be fully encoded.
Moreover, the current initial voxel encoding approach (HardSimpleVFE~\cite{yan2018second}) averages raw LiDAR points within each voxel.
It implies that inherent geometric information in the cell may be suppressed because information of point distribution in the voxel would be disregarded.
% This aspect is particularly crucial in our approach. Unlike previous backbones that rely on dense feature refinement, our LiDAR backbone, which solely encodes sparse features, heavily relies on the information within the raw point distribution.
To alleviate this problem, We present a simple modification for the initial voxel encoding by adding several statistics such as the standard deviation and the count of points within each cell into the average of them. We call this modified voxel feature encoding approach as mVFE.
This simple modification with a marginal overhead allows the raw voxel features to carry rich fine-level geometric information, consequently improving 3D object detection accuracy (Table \ref{tbl:mVFE}).

Specifically, original HardSimpleVFE~\cite{yan2018second} receives LiDAR points in each voxel, and each voxel is represented by five values (avg\_x, avg\_y, avg\_z, avg\_intensity, avg\_offset), where avg means an average value, intensity means the intensity of a LiDAR sensor, and offset means an temporal offset among multiple sweeps. In our mVFE, the representation is modified to have 11 values (avg\_x, avg\_y, avg\_z, avg\_intensity, avg\_offset, std\_x, std\_y, std\_z, std\_intensity, std\_offset, n\_points), where std means a standard deviation value, and n\_points means the number of points in the cell normalized to 0\(\sim\)1.

\subsection{Additional Architecture Detail}

Regarding the Transformer decoder, we use six transformer decoder layers, each of which consists of a self-attention operation, a cross-attention operation, and a feed-forward network. The number of queries used in our model is 900. The result of final transformer decoder layer is fed into our prediction heads, and the heads predict the center, scale, rotation, velocity, and class of each bounding cuboid.

Regarding our deep fusion module, we use four DSVT~\cite{wang2023dsvt} block, each of which consists of four set attention layers (along\_x, x\_shift, along\_y, y\_shift). The specific hyper-parameters are set\_info ([72, 4]), window\_shape ([24, 24, 11]), hybrid\_factor ([2, 2, 1]), and shifts\_list ([[0, 0, 0], [12, 12, 0]]).

% Regarding our feature elimination module, 

When we apply the deep fusion module to our model, we increase the target number of channels of our sparse encoder twice (128 to 256) because CMT also use 256 channels for BEV feature refinement as well as most recent models such as DAL~\cite{huang2023detecting} also use 256 channels for sparse encoding. We empirically found that 256 channels is minimally required for effective feature encoding.

%%%--------------------------------------------------------------------------------
\begin{table}[t]
\centering
\scalebox{0.92}{
\begin{tabular}{ c | c c }
\toprule
Model & mAP & NDS\\
\toprule
SparseVoxFormer-base w/o mVFE & 70.7 & 72.9 \\
SparseVoxFormer-base & 70.8 & 73.2 \\
\bottomrule
\end{tabular}
}
\vspace{-6pt}
\caption{Effect of using mVFE.}
\label{tbl:mVFE}
\vspace{-4pt}
\end{table}
%%%--------------------------------------------------------------------------------

\subsection{Additional Training Detail}
As we base our implementation on the source code of CMT~\cite{yan2023cross}, we follow many of the training details outlined in CMT.
Specifically, we employ ground-truth sampling during the training phase for the first 15 out of a total of 20 epochs. This is a form of curriculum learning. In the ground-truth sampling, instances containing fewer than 5 LiDAR points are excluded, \ie, those instances that belong to long-range and are difficult to detect. Also, additional filtering by difficulty and class grouping are applied.

For query denoising~\cite{li2022dn}, we add auxiliary queries using the center coordinates of ground-truth cuboids, but only during the training phase. Specifically, we introduce noise to the coordinates and use their positional embeddings as the initial transformer queries. A loss function for query denoising ensures that the output coordinates corresponding to the noisy auxiliary queries align with the ground-truth coordinates. To avoid trivial shortcuts, we separate the gradient groups for normal queries and the auxiliary queries, and prevent the gradients from the outputs of auxiliary queries from influencing other queries, as done in \cite{li2022dn,yan2023cross}.

We use a pretrained image backbone but train a LiDAR backbone from scratch. Regarding the image backbone, additional learning rate decays are applied for fine-tuning (0.01 for image backbone and 0.1 for image neck).
We use a cyclic learning rate policy with the initial learning rate of 0.0001, but modify the target ratio of (4, 0.0001) from (8, 0.0001) in the original CMT.
% The remaining loss weights are same with the those of CMT. 
We follow the training details of CMT for remaining factors (\eg the batch size is 16). Our training time is similar to that of CMT (about 2.5 days with eight A100 GPUs).

%%%--------------------------------------------------------------------------------
\begin{figure}[t]
\begin{center}
\includegraphics [width=0.80\linewidth] {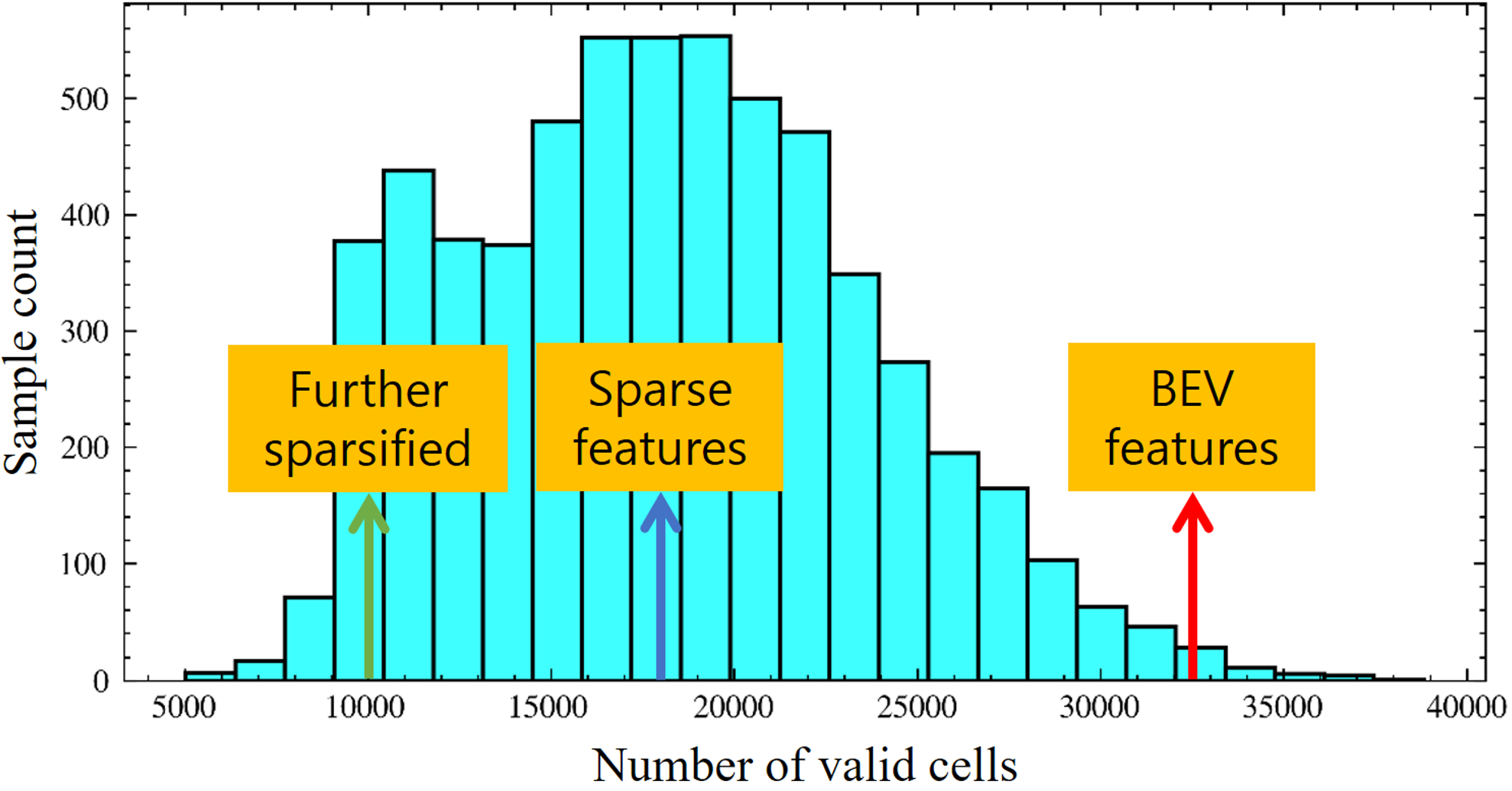}
\end{center}
\vspace{-8pt}
  \caption{Histogram of valid cell counts per LiDAR sample (10 sweeps) in the nuScenes train set, implying the distribution of the count of valid voxel features with the voxel resolution of \(180\times180\times11\). The red arrow denotes the number of cells for BEV feature map with the resolution of \(180\times180\). The average value (blue arrow) for the sparse voxel features is much smaller than the number of BEV features (red arrow), which is further reduced by our additional feature sparsification (green arrow).
  % \son{need the average value.}
}
\label{fig:histogram}
\vspace{-6pt}
\end{figure}
%%%--------------------------------------------------------------------------------

%%%--------------------------------------------------------------------------------
\begin{table}[t]
\centering
\scalebox{0.92}{
\begin{tabular}{ c | c | c c }
\toprule
&Dataset & ~~\# of tokens~~ & Feat. Resolution\\
\toprule
BEV&- & 32,400 & $180\times180\times1$ \\
\midrule
\multirow{3}{*}{Sparse}&Nuscenes & 18,000 & $180\times180\times11$ \\
&Waymo & 9,500 & $180\times180\times11$\\
&Argoverse2 & 3,000 & $180\times180\times11$ \\
\bottomrule
\end{tabular}
}
\vspace{-6pt}
\caption{Comparison of the number of tokens for BEV features and sparse voxel features according to various LiDAR sensors. Although the sparse 3D voxel features ($180\times180\times11$) from all the LiDAR sensors contain geometric information at a higher resolution than that of the BEV features ($180\times180$), the number of valid Transformer tokens is much smaller than that in the BEV features. }
\label{tbl:statistic}
\vspace{-4pt}
\end{table}
%%%--------------------------------------------------------------------------------

%%%--------------------------------------------------------------------------------
\begin{figure*}[t]
\begin{center}
\includegraphics [width=0.90\linewidth] {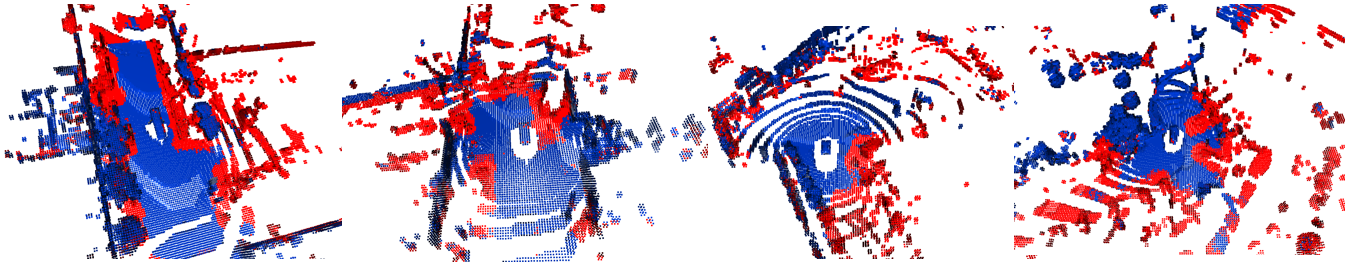}
\end{center}
\vspace{-8pt}
  \caption{Visualization of sparse features for several scenes. Blue points denote eliminated features by our feature elimination scheme, which is redundant for 3D object detection.
  % \son{need the average value.}
}
\label{fig:feature_elimination}
\vspace{-8pt}
\end{figure*}
%%%--------------------------------------------------------------------------------

\subsection{Additional Evaluation Detail}
In the analysis of computational costs, we measure Flops of multi-modal 3D object detectors.
To the best of our knowledge, there has been no well-established approach to calculate the computational complexity of multi-modal 3D object detector and models with sparse data and sparse operations to our best knowledge. Therefore, we manually computed them. Specifically, we separately inference LiDAR-only-based and camera-only-based branches, and manually combine their values. Furthermore, the computational cost of sparse convolution-based layers can vary according to the sparsity of LiDAR data. We assume the number of valid intermediate features as 20,000 in the layers for simplicity.

%%%--------------------------------------------------------------------------------
\begin{figure*}[t]
\begin{center}
\includegraphics [width=0.95\linewidth] {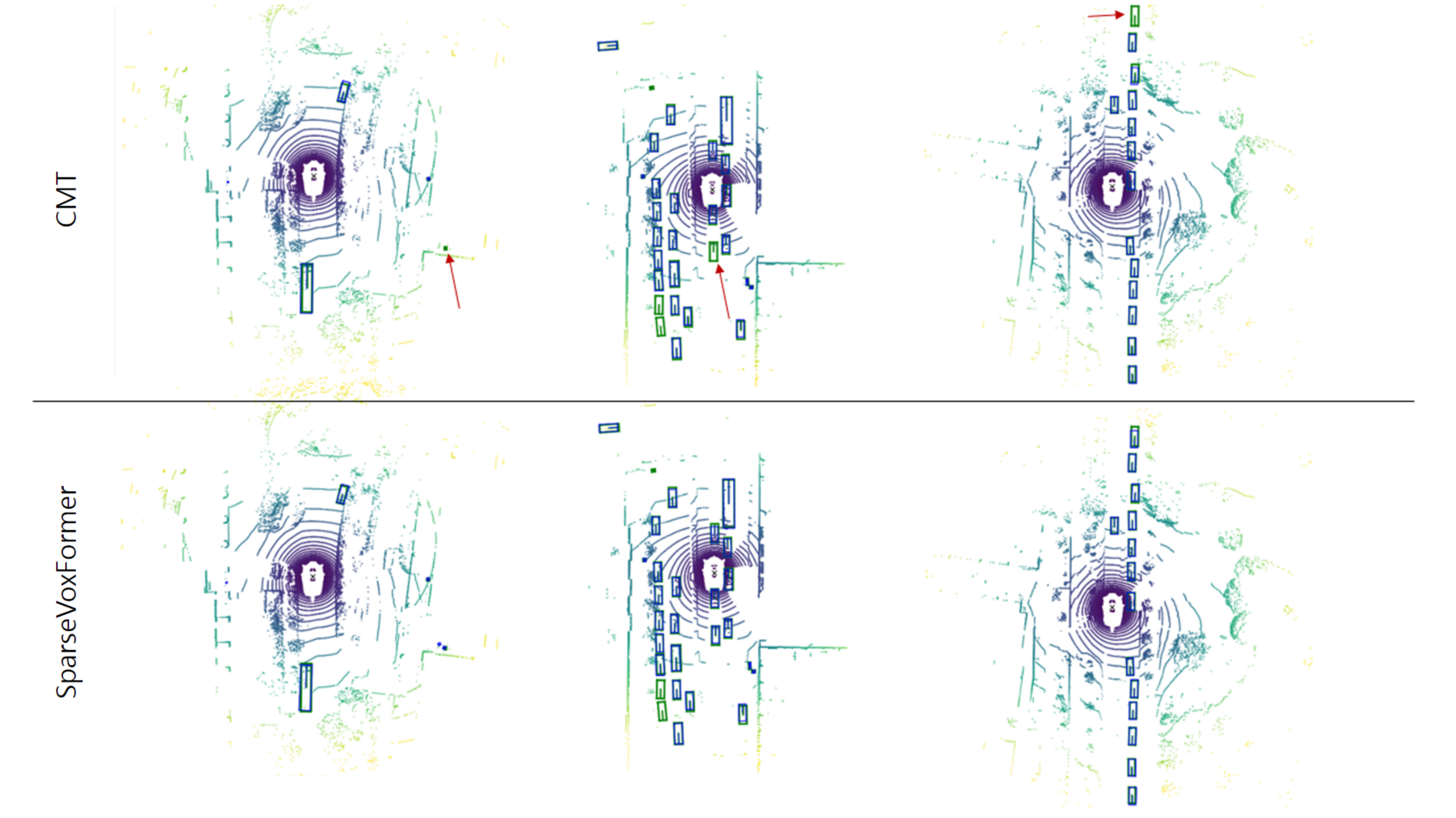}
\end{center}
\vspace{-16pt}
  \caption{Qualitative visualization of CMT~\cite{yan2023cross} and our SparseVoxFormer-L. A blue box denotes a prediction with a confidence score greater than 0.3, while a green box indicates a ground-truth bounding box. A red arrow highlights a noticeable difference in predictions between CMT and our SparseVoxFormer.
  % \son{should be modified}
}
\label{fig:vis_cmt_sp}
\vspace{-10pt}
\end{figure*}
%%%--------------------------------------------------------------------------------

\section{Additional Experiments}
% \son{may require a full training of Ours-L with mVFE for nuScene test set in our main paper. But, I am not sure that it will be positive for the reviews if we upload the new value in the supplemental material in this submission. There is another way to update the value in the camera-ready version if our paper gets accepted.}

% \son{How about test performance w/ TTA?}

\subsection{Deeper analysis of LiDAR statistics on Various LiDAR Sensors}
Our key motivation is based on the statistic that the number of valid transformer tokens in sparse 3D voxel features is significantly smaller than the number of 2D dense grids in the BEV space.
In this section, we confirm that this assumption generally holds across different datasets using various LiDAR sensors.
Fig.~\ref{fig:histogram} visualizes the LiDAR statistics from the nuScenes dataset~\cite{caesar2020nuscenes}, showing that sparse 3D voxel features yield fewer tokens than dense BEV features. However, in this case, the number of tokens for sparse features can vary due to differing sparsity in each scene. We emphasize again that our feature elimination scheme standardizes these varying numbers to a constant, such as 10,000, which is further reduced at the same time. 

We also analyzed LiDAR statistics from Argoverse2~\cite{Argoverse2} and Waymo open~\cite{Sun_2020_CVPR} datasets (Table. \ref{tbl:statistic}). We adopted common configurations of using the datasets shown in VoxelNeXT~\cite{chen2023voxenext} and DSVT~\cite{wang2023dsvt}. They use a single LiDAR sweep per scene compared to nuScenes' 10 sweeps, resulting in about 3,000 (Argoverse2) and 9,500 (Waymo) valid cells in average for $180\times180\times11$ voxels, indicating rather higher sparsity than 18,000 (nuScenes). 
Based on the statistics, we believe that our approach is highly efficient for 3D object detection using general LiDAR sensors.

\subsection{Visualization of Feature Elimination}
Fig.~\ref{fig:feature_elimination} visualizes the voxels removed through our feature elimination scheme. As shown in the examples, the eliminated features are primarily placed on backgrounds such as roads.

\subsection{Qualitative Results}
Fig.~\ref{fig:vis_cmt_sp} shows visual 3D object detection results from a top view. As demonstrated in the left and right examples, our approach can detect long-range small objects. In the middle example, it shows better handling of an occluded object compared with CMT~\cite{yan2023cross}.

\section{Discussion of Limitation and Future Work}
Our voxel features are derived from voxels that contain at least one point. Furthermore, our multi-modal fusion approach only combines LiDAR voxel features with their corresponding image features, resulting in sparse data. Therefore, our approach may be unable to handle any region without LiDAR points. Nevertheless, we demonstrate that our approach effectively detects long-range objects with few LiDAR points, compared to BEV-based approaches, as shown in Table 6 in the main paper. Additionally, the sparsity of LiDAR data depends on the hardware specification of the LiDAR sensor, meaning the efficiency of our model could vary. However, we believe that our additional feature elimination scheme provides a viable solution to this limitation.

Despite our achievements, we believe there are further future directions that could enhance this innovative approach, as we have presented several sparse feature-specific designs. We have focused on transformer keys in this work, but exploring new query designs like iterative query refinement could be interesting. Other research areas could include using sparse features for different 3D perception tasks, not just 3D object detection.

\paragraph{Potential negative societal impact}
3D object detection is a crucial task for autonomous driving. In the current paradigm, the planning of autonomous vehicles often relies on the performance of 3D object detection. As our approach enhances the performance of 3D object detection, it can be utilized to improve the overall performance of autonomous driving. However, 3D object detection models may still produce errors when encountering corner cases, subsequently posing a potential risk of influencing incorrect decisions in autonomous vehicles.

\end{document}